\documentclass[conference]{IEEEtran}
    \IEEEoverridecommandlockouts
    \usepackage{cite}
    \usepackage{amsmath,amssymb,amsfonts}
    \usepackage{graphicx}
    \usepackage{textcomp}
    \usepackage{float}
    \usepackage[table]{xcolor}
    \usepackage[T1]{fontenc}
    \begin{document}
    
    \title{Benchmarking 3D Reconstruction for Under-Ice Robotic Perception in Arctic Environments}
    
    \author{
        \IEEEauthorblockN{
            Klára Orbán\textsuperscript{*1}, 
            Ashwin Nedungadi\textsuperscript{*2}, 
            Ana-Ștefania Kogălniceanu \textsuperscript{1}, 
            Bianca-Marcela Sava\textsuperscript{1},
        }
        \IEEEauthorblockN{
            Bastian Cieply\textsuperscript{2}, 
            Raul-Alexandru Simon\textsuperscript{1}, 
            Kuderna-Iulian Bența\textsuperscript{1},
            Stefan Lüdtke\textsuperscript{2}
        }
        \IEEEauthorblockA{
            \textsuperscript{1}Faculty of Mathematics \& Computer Science, Babeș-Bolyai University, Cluj-Napoca, Romania\\
            \{klara.orban, ana.kogalniceanu, bianca.sava, raul.simon\}@stud.ubbcluj.ro,
            kuderna.benta@ubbcluj.ro
        }
        \IEEEauthorblockA{
            \textsuperscript{2}Institute of Visual \& Analytic Computing, University of Rostock, Germany\\
            \{ashwin.nedungadi, bastian.cieply, stefan.luedtke\}@uni-rostock.de
        }
        \thanks{\textsuperscript{*}Equal Contribution.}
    }
    
    \maketitle
    
    \begin{abstract}
    As the Arctic transitions toward a seasonally ice-free state, capturing the complex morphology of submerged sea ice is crucial for advancing models of ocean-ice-atmosphere heat exchange and monitoring under-ice habitats. 
    
    Currently, sub-surface ice mapping is predominantly achieved through acoustic sensors such as multibeam sonar. While sonar provides robust large-scale geometry, it lacks spectral and high-resolution textural information required to record fine morphological features - a critical limitation in under-ice environments, where sparse visual features and degraded sensing conditions challenge the capture of detailed structural and optical surface properties.
    
    We present a benchmark for vision-based under-ice 3D reconstruction using real-world data from the MOSAiC expedition and synthetic data generated in an Unreal Engine-based underwater robotics simulator. The benchmark evaluates Structure-From-Motion (SfM) and monocular depth estimation, with quantitative comparison enabled through cross-modal registration to upward-looking multibeam sonar.
    \end{abstract}
    
    
    \section{Introduction}
    The complex morphology of the underside of sea ice poses significant challenges for robotic perception and mapping in polar environments. Under-ice exploration relies on onboard sensing to operate in regions inaccessible to direct human observation, making accurate geometric representations of surrounding ice structures critical for navigation and situational awareness \cite{wadhams-2006}. Features such as ridges, keels, and melt-induced cavities create highly irregular surfaces that complicate both visual and acoustic sensing. High-resolution surface mapping  of the ice underside therefore plays a key role in developing and evaluating computer vision and sensor fusion pipelines, improving localization and obstacle awareness for under-ice robotic systems, and enabling realistic testing of autonomous navigation in these challenging environments.
    
    Traditionally, acoustic sensing has been used to characterize subsurface ice textures \cite{sonar_mapping}. The predominant modality for obtaining large-scale geometry has involved upward-facing multibeam sonars operated from Autonomous Underwater Vehicles (AUVs) or Remotely Operated Vehicles (ROVs). 
    
    While acoustic sensing provides robust geometric data, it is inherently insensitive to color variations and unable to capture optical information such as light transmittance, albedo or surface texture. Visual photogrammetry, specifically Structure-from-Motion (SfM) \cite{agisoft_metashape_2023}, offers a framework to address this information gap. However, the subsurface ice remains a significant challenge to computer vision approaches because of extreme light attenuation, backscatter from suspended scatterers, and the sparse feature texture of ice surfaces, which are often repetitive in nature, leading to significant geometric instability.
    
    In this work, we introduce a benchmark for under-ice robotic perception based on data collected during the MOSAiC expedition \cite{mosaic_ack}. We evaluate vision-based 3D reconstruction methods by comparing Structure-from-Motion (SfM) reconstructions generated using Agisoft Metashape \cite{agisoft_metashape_2023} and monocular depth estimation using Depth Anything V3 (DA3) \cite{lin2025depth3recoveringvisual} against coincident upward-looking multibeam sonar measurements and analyze their domain-dependent behaviour across real and simulated under-ice environments.
    
    The main difficulty in enabling such a comparison is the lack of direct spatial correspondence between acoustic and visual data, as the two modalities differ in sensing principle, sampling density and observable coverage. To address this, we establish a common evaluation domain that enables consistent comparison between the two sensing modalities.
    
    Collecting additional real-world data is logistically impractical, considering that Arctic expeditions are infrequent, costly, and weather-constrained. A simulated environment is a highly controlled setting where lighting, turbidity, ROV paths, and sensor settings can be precisely defined and consistently reproduced. Utilizing the geometry data from real-world MOSAiC sonar surveys ensures that our simulated environment is realistic and also fully controllable.
    
    In summary, the contributions of this paper are:
    \begin{itemize}
        \item a benchmark for evaluating vision-based 3D reconstruction methods against spatially aligned under-ice multibeam sonar data,
        \item a cross-modal alignment procedure that defines a shared sensing region for quantitative comparison between optical and acoustic reconstructions,
        \item a simulation pipeline in HoloOcean based on sonar-derived ice geometry for generating controlled multimodal under-ice datasets.
    \end{itemize}
    
    \section{Related Work}
    \subsection{Acoustic and Optical Sensing}
    Multibeam echosounders are a primary instrument for sub-ice topographic mapping, measuring the two-way travel time and angle of acoustic pulses emitted across a wide swath to reconstruct ice draft geometry. By combining these range measurements with vehicle navigation data, dense three-dimensional representations of the ice underside can be generated over large spatial extents \cite{anhaus-2024}. Due to their robustness to low visibility and light-independent operation, acoustic systems are widely regarded as the standard tool for large-scale geometric reconstruction in under-ice environments.
    Despite these advantages, multibeam-derived reconstructions are subject to several sources of uncertainty. In particular, inaccuracies in vehicle pose estimation, such as roll miscalibration or lateral positioning drift, can introduce geometric distortions in the reconstructed surface. Furthermore, acoustic methods are inherently limited in that they do not capture optical properties such as transmittance, albedo, or biological pigmentation, which are essential for applications such as radiative transfer modeling.
    Visual photogrammetry, specifically Structure-from-Motion (SfM), offers a framework to complement these limitations by reconstructing geometry while preserving optical appearance. However, when SfM is used to study the ice-water interface, there are particular environmental challenges, such as flat port distortion \cite{cimoli-2019}, pincushion distortion, and those related to camera network geometry. While techniques such as 3D-informed image formation models have the capability to recover surface reflectance \cite{bryson-2015}, SfM reconstructions remain highly sensitive to feature sparsity and light scattering, characteristics specific to under-ice surveys \cite{burns-2015, qiao-2019}.
    
    \subsection{Monocular Depth Estimation}
    Monocular depth estimation has emerged as a promising alternative to traditional multi-view reconstruction methods, aiming to recover scene geometry from single images using learned priors. These approaches are  attractive in underwater and under-ice environments, where multi-view consistency is difficult to achieve due to limited visibility, sparse viewpoints and unstable camera motion. Typical methods range from self-supervised techniques that exploit temporal consistency in image sequences to supervised and foundation-model-based approaches trained on large-scale datasets.
    However, deploying these methods in underwater environments presents significant challenges. Self-supervised techniques rely heavily on photometric reprojection loss, which assumes brightness constancy between frames. This assumption is broken in underwater settings due to backscatter, light attenuation and dynamic illumination changes, leading to degraded performance \cite{amitai-2022}. While self-supervised Vision Foundation Models, such as Depth Anything V3, have shown improved results with synthetic fine-tuning \cite{cai-2025}, none have been tested on upward-facing subsurface-ice imagery, which represents an extreme out-of-distribution case to the photometric priors learned from terrestrial datasets. The significant scattering and lack of distinctive visual features in underwater environments beneath Arctic ice further limit the reliability of purely learning-based depth estimation models.
    
    \subsection{Simulation and Data Benchmarking}
    Simulation is essential in addressing the lack of annotated data in underwater environments. While OceanSim \cite{song-2025} and UNav-Sim \cite{amer-2023} are highly effective in sensor modeling and ROS2 support, they are often based on procedural or artificially designed environments. The MIMIR-UW dataset \cite{alvareztunon-2023} proves the effectiveness of artificially generated data in navigation tasks, even recognizing the sim-to-real gap that arises due to the inability of artificially generated images to mimic real-world scattering effects in water.
    
    \section{Datasets}
    \label{sec:datasets}
    The primary dataset used to answer the research questions was collected during the Multidisciplinary Drifting Observatory for the Study of Arctic Climate (MOSAiC) expedition \cite{Anhaus2025, Anhaus_2025_2}. By providing both under-ice topography and image data gathered with a multibeam sonar system and a high-definition camera rigidly co-mounted on the ROV’s upper platform in an upward-looking orientation, this dataset supports the reconstruction experiments presented in this work. We prioritized and primarily worked with survey PS122/3\_39-20, conducted on May 5, 2020, during the MOSAiC expedition. It provided the most visually consistent and geometrically complete overlap between acoustic and optical sensor streams required for an accurate comparison.
    
    The acoustic measurements were acquired using an Imagenex DT101 multibeam sonar operating at 240 kHz. The system emits 480 beams simultaneously with a swath width of 120° across-track and 3° along-track, resulting in an effective beam width of approximately 0.75°and an angular resolution of 0.25°. During survey operations the ROV was typically operated at a depth of approximately 20 m below the ice underside for standard surveys, while higher-resolution surveys were conducted at a depth of 10 m.
    
    Survey trajectories were flown in a grid or sweep pattern in order to ensure sufficient spatial overlap between adjacent passes. Horizontal spacing between lines ranged between 20 m and 25 m, providing approximately 30\% overlap of the area scanned by neighboring swaths. The sonar operated at an average ping rate of approximately 9.37 Hz. Standard survey processing resulted in horizontal grid resolutions of approximately 0.5 m, while higher-resolution products achieved lateral resolutions of approximately 0.1 m.
    
    \subsubsection{Acoustic Data Processing}
    The raw sonar point clouds contain spurious measurements caused by acoustic noise, multi-path reflections, and surface scattering effects typical of imaging sonar systems. Automated statistical filtering methods, such as the Statistical Outlier Removal (SOR) filter implemented in CloudCompare, were initially evaluated but proved overly aggressive, removing valid structural measurements along with noise.
    
    A geometric preprocessing strategy was therefore adopted. Outliers were removed using spatial clipping based on inspection of the point cloud distribution. Points exhibiting large deviations from the main surface along the vertical axis were removed using axis-aligned clipping, while points outside the spatial extent of the structure were eliminated using planar clipping in the horizontal plane. This approach removes clearly erroneous measurements while preserving the geometric characteristics of the ice underside.
    
    While multibeam sonar has inherent uncertainties in vehicle position, its direct physical range measurements remain unaffected by the lighting and texture-related difficulties that degrade optical sensing, and its high-frequency acoustic measurements at 240kHz provides a reference significantly more stable than the decimeter to meter-level errors typical of Structure-from-Motion (SfM) in under-ice conditions. Consequently, sonar is chosen as the operational ground truth for this benchmark, representing the best available geometric baseline in this environment.
    
    \subsubsection{Optical Data Processing}
    The dataset used in this study contains two optical data types: (i) forward-looking video sequences recorded with a Bowtech Surveyor WAHD camera (1920×1080), and (ii) upward-looking still images captured with an Imenco Tiger Shark camera (4320×3240). Video frames were used in auxiliary reconstruction experiments but not in the primary quantitative results.
    
    The video data contains navigation overlays and watermarks, which interfere with feature-based reconstruction by being incorrectly detected as visual features. To address this, several preprocessing strategies were evaluated, including masking overlay regions, infilling using surrounding pixels, and cropping regions with persistent artifacts.
    
    The still image dataset did not exhibit these issues and was therefore used as the main input for image-based reconstruction. Image enhancement, including sharpening and haze reduction \cite{woteqEnhanceUnderwater}, was applied as domain alignment, since models such as DA3 are trained on terrestrial datasets and perform under out-of-distribution conditions in underwater imagery \cite{wu2025stereoadapteradaptingstereodepth}.
    
    No intrinsic camera parameters (e.g., focal length or lens distortion) were available for either camera system.
    
    \section{Methods}
    \subsection{Geometric Reconstruction}
    \label{sec:geometric-rec}
    \subsubsection{Structure-from-Motion: Feature Matching in Sparse Environments}
    Three-dimensional reconstruction of the ice underside was performed using Agisoft Metashape\cite{agisoft_metashape_2023}, a commercial SfM photogrammetry package. In preliminary experiments, standard COLMAP failed to produce stable camera alignments on the MOSAiC imagery, consistent with these known limitations. Metashape's feature detection and matching implementation demonstrated greater robustness under these conditions, recovering partial alignments where COLMAP produced no output.
    \begin{figure}[htbp]
    \includegraphics[width=\columnwidth]{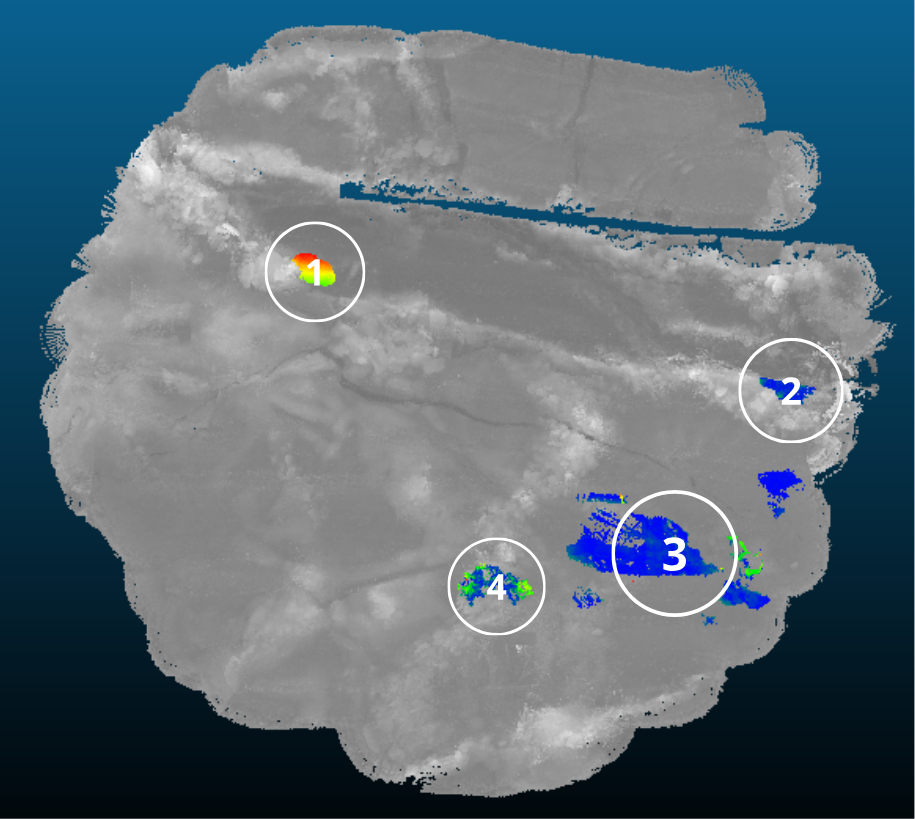}
    \caption{Cross-modal alignment in Area 1: Composite overlay showing four-point registration between multibeam sonar geometry and coincident optical imagery.}
    \label{fig:area1}
    \end{figure}
    The reconstruction pipeline consisted of feature detection and matching, camera pose estimation, dense point cloud generation, and mesh reconstruction. ROV navigation coordinates from the MOSAiC survey were imported as reference positions to constrain camera alignment, reducing reliance on purely image-based pose estimation. As these coordinates are expressed relative to the start of the survey trajectory rather than an absolute reference frame, they serve to initialise alignment rather than replace it entirely.
    
    Reconstruction quality varied substantially across the survey trajectory. Stable results were obtained primarily in Area 1 (see Fig.~\ref{fig:area1}), where image coverage and feature visibility were most consistent, while higher turbidity and featureless ice regions produced incomplete or geometrically unstable reconstructions.
    
    \subsubsection{Monocular Depth Estimation}
\begin{figure}[htbp]
    \includegraphics[width=\columnwidth]{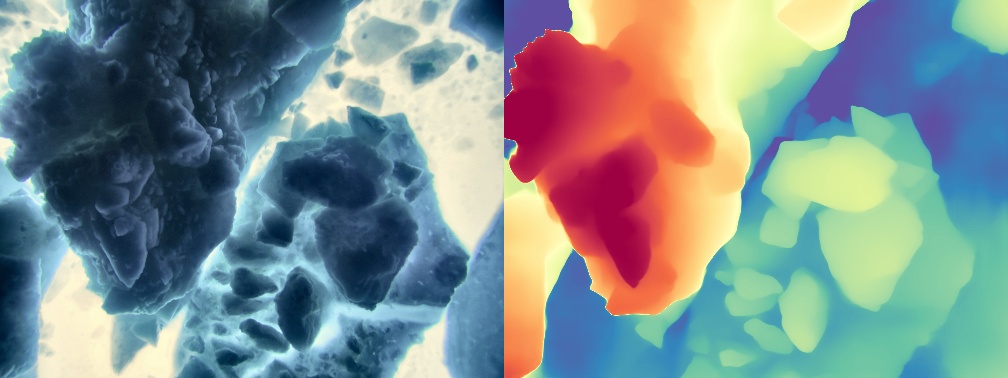}
    \caption{DA3 depth estimation: The original ice image (left) and the resulting depth map (right).}
    \label{fig:da3_fig}
    \end{figure}

    As an alternative to multi-view photogrammetry, monocular depth estimation approaches that infer scene depth from individual images using deep neural networks were also investigated. Depth Anything V3 (DA3)\cite{lin2025depth3recoveringvisual} was used to estimate per-image depth maps from the selected frames (as illustrated in Fig.~\ref{fig:da3_fig}). This enables 3D reconstruction from images where intrinsic and extrinsic parameters are not provided a priori, since the model estimates per-frame focal length and principal point as well as the rotation matrix and translation vector directly from the image content. The predicted camera parameters and depth maps are then used to create three-dimensional point clouds.
    
    Point clouds exhibiting surface continuity and absence of isolated outlier clusters were aligned to the sonar reference using the procedure described in the following section.
    
    \subsubsection{SfM-Sonar Alignment}
    A central challenge of this work is establishing consistent spatial correspondence between acoustic and visual reconstructions. Because the two sensing modalities differ in measurement principle, spatial sampling, and observable coverage, quantitative evaluation is only possible within regions jointly observed by both.
    
    To this end, evaluation was restricted to the spatial region observed by both sensing modalities. The same alignment and cropping procedure was applied whenever the reconstructed point cloud permitted stable cross-modal registration, enabling a consistent evaluation protocol across real and simulated datasets.
    
    The reconstruction (SfM or DA3) was aligned to the sonar point cloud, which served as the primary geometric reference due to its direct acoustic range measurements. An initial coarse alignment was performed manually to establish approximate spatial correspondence between the two point clouds. Minor adjustments, primarily along the vertical axis, were required to account for differences in sensing geometry and uncertainty inherent to photogrammetric reconstruction. Following this coarse alignment, the transformation between the point clouds was refined using the Iterative Closest Point (ICP) algorithm to improve geometric consistency between the two modalities.
    
    After alignment, the reconstructed point cloud was used to approximate the spatial footprint of the visual reconstruction. A slightly expanded bounding region around the reconstruction was used to crop the sonar data, defining a conservative overlap volume representing the portion of the ice underside within the camera's effective field of view. Quantitative comparisons between the sonar and reconstructions were therefore computed only within this overlapping region.
    
    This approach avoids penalizing the reconstruction for regions that were not observable by the camera system while preserving the sonar data as the geometric reference representation of the scene. Because the two sensing modalities rely on fundamentally different measurement principles—acoustic ranging for sonar and either photogrammetric triangulation (SfM) or learned depth inference (DA3), small discrepancies in point density and local surface representation are expected.
    
    This alignment procedure forms the foundation of the proposed benchmark, enabling consistent cross-modal evaluation across both real and simulated datasets.
    
    \subsection{Simulation and Synthetic Environment}
    The HoloOcean \cite{romrell2025previewholoocean20} framework mitigates the inherent noise and gaps present within the field-gathered MOSAiC dataset. Using the hardware-efficient rendering capabilities provided by Unreal Engine, we deployed an eight-motor remotely operated vehicle (ROV) platform within a controlled environment with customized camera and sonar configurations. In order to simulate the Arctic environment, an ice mesh  was imported based on actual sonar data \cite{anhaus-2024}, with custom materials used to visually simulate underwater scenes and approximate the conditions at the interface of ice and water. Additionally, water column parameters were calibrated to simulate real-world properties such as subsurface light scattering, with dense volumetric fog and non-uniform light attenuation to simulate the visual constraints within the underwater environment. The simulated ROV carries a synchronized RGB camera and a simulated multi-beam sonar, which match the sampling rate and geometry with respect to the MOSAiC hardware set. Using a predefined grid-based mission trajectory, we generated a multimodal synthetic dataset\cite{aznee2024arcticode} with precise geometric ground truth and high overlap between the simulated sensors to provide a rigorous baseline for isolating the specific failure modes for under-ice perception algorithms.
    
    \subsection{Experimental Procedure}
    To ensure reproducibility and clarity of the evaluation, we explicitly define the data selection, reconstruction workflow and evaluation protocol used in this study.
    \newline
    \textbf{MOSAiC Field Data:}
    Experiments were conducted on survey PS122/3\_39-20. Frames were manually selected based on illumination, turbidity, and viewpoint variation. Multiple spatial regions along the ROV trajectory were reconstructed using the selected frames. Among these, Area 1 (Fig.~\ref{fig:area1}), consisting of approximately 160 images, provided the most complete image coverage and viewpoint diversity. The remaining three regions, despite having a comparable number of frames, exhibited steeper viewing angles, reduced illumination, and increased turbidity. These factors resulted in weaker feature correspondences and reduced geometric stability.
    All reconstructions were processed using the pipeline described previously(Section~\ref{sec:geometric-rec}).
    \newline
    \textbf{Simulated Data (HoloOcean):}
    Synthetic data were generated using a grid-based ROV trajectory of approximately 90 seconds, from which image frames were extracted. The trajectory was chosen to maximize viewpoint diversity and coverage of larger ice structures.
    
    The simulated platform was equipped with a co-registered RGB camera and multibeam sonar matching the real system (Section~\ref{sec:datasets}). The environment was constructed from sonar-derived ice geometry, ensuring realistic morphology while enabling controlled variation of environmental conditions.
    
    Both SfM (Metashape) and DA3 were applied to the simulated dataset using the same reconstruction and alignment pipeline (Section~\ref{sec:geometric-rec}).
    
    \section{Experimental Results}
    \subsection{MOSAiC Field Data}
    In practice, SfM consistently produced spatially coherent reconstructions under real-world conditions, whereas DA3 remained unstable and only produced usable results in isolated cases.
    
    Stable SfM reconstructions were obtained primarily in Area~\ref{fig:area1}, where overlapping images and sufficient feature visibility allowed for robust camera pose estimation. In other areas, increased turbidity and feature-poor ice surfaces led to partial alignments and geometrically unstable results, consistent with known limitations of feature-based photogrammetry in optically degraded underwater settings.
    
\begin{figure}[htbp]
    \includegraphics[width=\columnwidth]{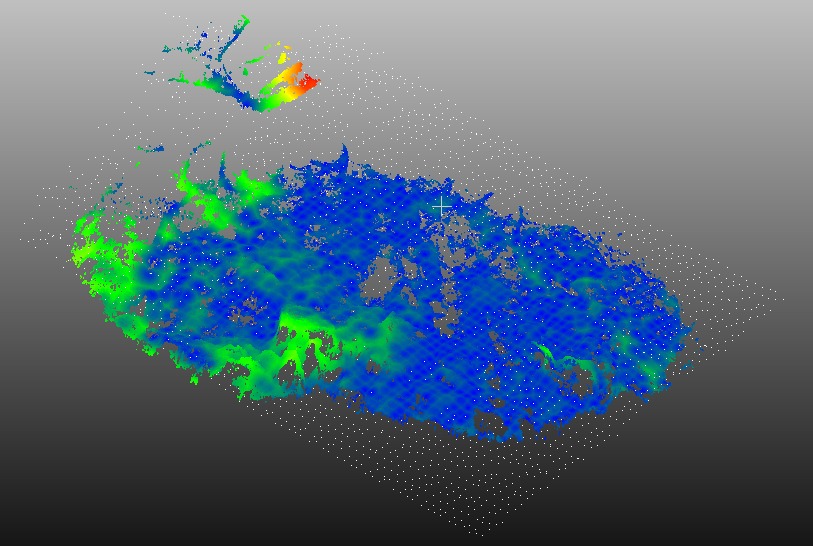}
    \caption{CloudCompare visualization of the alignment between the SfM reconstruction and the sonar reference. The sonar slice is shown in white, while the SfM point cloud is colored by C2C (cloud-to-cloud) distance (m). Blue and green indicate good alignment, while yellow-red regions indicate larger deviations.}
    \label{fig:cc_vis}
    \end{figure}

    After cross-modal alignment using iterative closest point (ICP) (visualized in Fig.~\ref{fig:cc_vis}), the SfM reconstruction aligns with the sonar reference with RMSE of 0.5525\,m and MAE of 0.4508\,m relative to the sonar reference. Point-to-point errors are small relative to the overall spatial extent, suggesting that the large-scale relief and primary morphological structures are recovered in regions with sufficient image coverage. However, surface distance measures indicate localized geometric deviations, particularly in areas affected by turbidity, floating debris, and limited feature availability.
    
\begin{figure}[htbp]
    \centering
    \includegraphics[width=0.5\columnwidth]{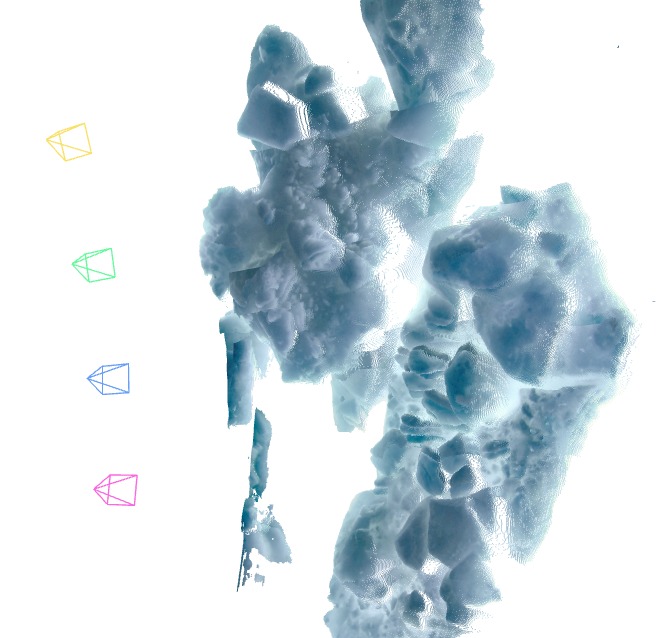}
    \caption{Example of reconstruction with DA3 inference using four enhanced images and estimated camera poses.}
    \label{fig:da3_res}
    \end{figure}

    In contrast, DA3 reconstructions of the MOSAiC dataset were generally fragmented and lacked consistent geometric structure (see Fig.~\ref{fig:da3_res}). This lack of identifiable geometric correspondences between the optical and sonar data prevented reliable alignment in most cases. In Area 1, where visibility and feature content were higher, partial reconstructions could be obtained, but their geometric accuracy remained inferior to SfM.
    \begin{table}[H]
    \centering
    \caption{Reconstruction error metrics on MOSAiC field data. All distance-based metrics are reported in meters.}
    \label{tab:mosaic_metrics}
    \scriptsize
    \setlength{\tabcolsep}{4pt}
    \begin{tabular}{l cc}
    \hline
    \textbf{Metric} & \cellcolor{green!25}\textbf{SfM (Metashape)} & \cellcolor{red!25}\textbf{DA3 (Monocular)} \\
    \hline
    RMSE (m)      & 0.5525 & 1.2103 \\
    MAE (m)       & 0.4508 & 0.8569 \\
    ARE (-)       & 0.0107 & 0.0204 \\
    W-RMSE\footnotemark (m)    & 0.5761 & 0.9705 \\
    Chamfer (m)   & 1.7036 & 3.0413 \\
    Hausdorff (m) & 9.0871 & 13.4949 \\
    DA-Ch. \cite{wu2021densityawarechamferdistancecomprehensive} (m)    & 1.6984 & 2.5935 \\
    \hline
    \end{tabular}
    \end{table}
    \footnotetext{Weighted by point cloud density for a representative global error estimate.}
    \subsection{HoloOcean Simulated Data}
    In the simulated environment, reconstruction performance differs significantly from the MOSAiC dataset.
    
    SfM fails to produce stable results, with large geometric distortions and inconsistent structure across the reconstructed regions. This is attributed to low-contrast surface textures and uniform lighting models, which reduce feature distinctiveness and degrade feature matching.
    
\begin{figure}[htbp]
    \includegraphics[width=\columnwidth]{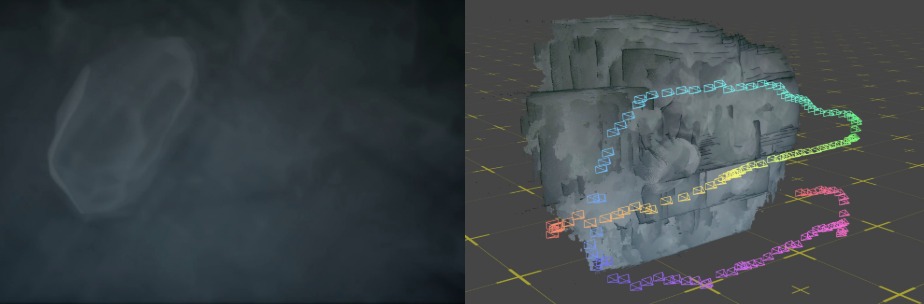}
    \caption{Comparison of the simulated environment and the resulting DA3 reconstruction: (left) the HoloOcean synthetic underwater viewport and (right) the generated 3D point cloud using the simulated camera trajectory.}
    \label{fig:sim_compare}
    \end{figure}

    Conversely, DA3 produces spatially coherent reconstructions that can be aligned with the sonar reference (visualized in Fig.~\ref{fig:sim_compare}). This behavior is supported by the controlled imaging conditions, including stable illumination and consistent surface appearance, which enable more reliable depth prediction.
    
    \begin{table}[h]
    \centering
    \caption{Reconstruction error metrics on HoloOcean simulated data. All distance-based metrics are reported in meters.}
    \label{tab:holo_metrics}
    \scriptsize
    \setlength{\tabcolsep}{4pt}
    \begin{tabular}{l cc}
    \hline
    \textbf{Metric} & \cellcolor{red!25}\textbf{SfM (Metashape)} & \cellcolor{green!25}\textbf{DA3 (Monocular)} \\
    \hline
    RMSE (m)      & 10.2149 & 0.3176 \\
    MAE (m)       & 7.8551  & 0.2414 \\
    ARE (-)       & 0.3145  & 0.0097 \\
    W-RMSE (m)    & 10.3863 & 0.2917 \\
    Chamfer (m)   & 10.1127 & 0.7839 \\
    Hausdorff (m) & 79.5502 & 4.0488 \\
    DA-Ch.\cite{wu2021densityawarechamferdistancecomprehensive} (m)    & 10.1610 & 0.7563 \\
    \hline
    \end{tabular}
    \end{table}
    
    \section{Discussion}
    
    Results reveal a domain dependence. On real MOSAiC data, only SfM produced geometrically meaningful reconstructions, whereas DA3 fails to align reliably with the sonar reference. In simulation, the opposite trend was observed: DA3 produced coherent reconstructions with lower geometric error, while SfM struggled under the rendered image conditions. When using real-world data from MOSAiC, SfM achieved significantly better results for geometric errors (RMSE: 0.55 m), whereas the DA3 algorithm could not align itself with the sonar reference in the majority of tested regions (RMSE: 1.21 m). Although specific instances of improved DA3 results were observed under low turbidity, the difference was insignificant, and still did not outperform SfM (RMSE: 1.21 m vs. 0.55 m; Chamfer: 3.07 m vs. 1.70 m).
    
\begin{figure}[htbp]
    \centering
    \includegraphics[width=1\columnwidth]{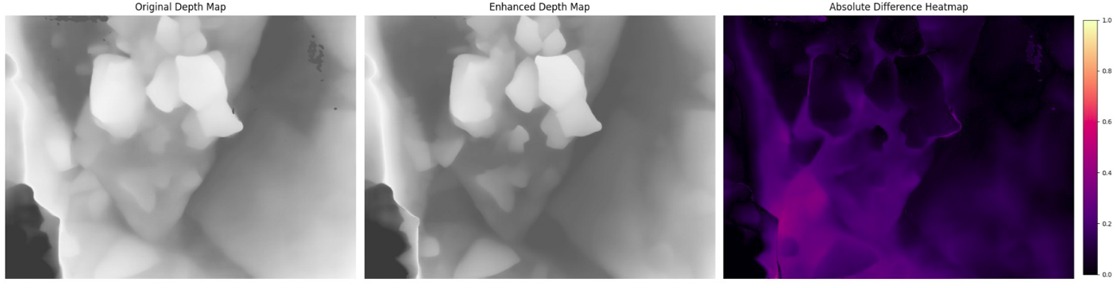}
    \caption{Depth map enhancement and absolute difference heatmap showing localized refinements. (Per-pixel absolute difference (normalized) between the original and enhanced grayscale depth maps).}
    \label{fig:depth_dif}
    \end{figure}

    In contrast, when using simulated data, DA3 delivered significantly better results with lower geometric error (RMSE: 0.32 m), while SfM delivered substantially worse results (RMSE: 10.21 m). On real MOSAiC data, DA3 failed to perform well despite the improvements made by photometric enhancement. As shown in Fig. \ref{fig:depth_dif}, preprocessing can enhance the local structure alignment, but cannot overcome the gap between simulation and reality. 
    \newline
    Real under-ice data represents a severe out-of-distribution challenge that breaks the learned spatial priors of foundation models like DA3, a sim-to-real limitation similarly observed in stereo-based underwater networks \cite{wu2025stereoadapteradaptingstereodepth}. To begin addressing this gap, our HoloOcean extension provides a controlled environment for generating multimodal datasets intended to fine-tune vision-based models for these harsh domains. Ultimately, overcoming these perception limits requires fusing optical sensing which captures fine-scale photometric details unresolved by acoustic footprints with the globally stable geometric reference of multibeam sonar. Advancing high-fidelity mapping in extreme subsea environments therefore depends on tightly integrating textural fidelity of computer vision with the structural robustness of acoustic baselines.
    \subsection{Dataset Scarcity}
    The scarcity of suitable under-ice optical datasets represents the most fundamental limitation facing this line of research. Existing surveys were designed primarily around acoustic sensing and scientific observation rather than photogrammetric reconstruction. As a result, available image collections are characterised by sparse spatial coverage, inconsistent inter-frame overlap, and ROV trajectories that were not optimized for systematic scene coverage. The closest precedent is the proof-of-concept deployment\cite{cimoli-2019}, which covered short transects over landfast ice under controlled sled conditions, which is far removed from the ROV-based, open-water survey geometry required here. Larger campaigns such as MOSAiC prioritised acoustic mapping and bio-optical measurements, with optical imagery collected as a secondary output rather than a primary data product. The result is that the type of dense, systematically overlapping image sequences that SfM requires has never been a collection priority in under-ice survey design.
    
    \subsection{Environmental Constraints}
    Beyond data availability, the under-ice environment itself introduces physical constraints that no improvement in data collection can fully overcome. While the ice underside varies significantly in geometric relief, from flat congelation ice to heavily deformed pressure ice, it remains visually featureless across both surface types, providing insufficient texture for feature-matching pipelines to establish reliable correspondences. Abrupt changes in the ice draft introduce a large inter-frame geometric displacement that, in the absence of distinctive surface features, produces reconstruction failures predominantly where the geometry is most variable.
    This is further compounded by the optical properties of the water column. Turbidity driven by suspended particles and biological matter introduces spatially uneven scattering, degrading image sharpness, while active water currents continuously shift the optical properties of the water column between consecutive frames, violating the scene stationarity assumption at the core of SfM. Together, these factors represent intrinsic properties of the Arctic under-ice environment rather than limitations that can be resolved through improved hardware or software alone.
    
    \section{Conclusion}
    We presented a benchmarking framework for vision-based under-ice 3D reconstruction using real-world MOSAiC data and HoloOcean synthetic environments, with multibeam sonar serving as the geometric ground truth. These findings emphasize the complementary nature of acoustic and optical sensing. Sonar provides a stable geometric reference, while vision-based methods capture fine-scale surface detail. The severe sim-to-real gap observed in foundation models highlights the necessity of domain adaptation for harsh environments, a challenge our simulation pipeline addresses through controlled data generation. Future research will focus on bridging this gap by expanding the simulator to encompass a wider diversity of ice morphologies. By leveraging this synthetic data, we intend to fine-tune vision foundation models against the photometric and scattering effects inherent to under-ice environments. Ultimately, we aim to translate this work into a computationally efficient, real-time perception pipeline.
    
    \section*{Acknowledgment}
    The authors express thanks to the MOSAiC expedition team for facilitating the collection of the under-ice data used in the research. The authors express thanks to the Alfred Wegener Institute (AWI) for the support in the data collection process. Special thanks go to Florian Zimmer for the dataset and technical support throughout the research process.
    
    Data used in this manuscript was produced as part of the international 
    Multidisciplinary drifting Observatory for the Study of the Arctic Climate 
    (MOSAiC) with the tag MOSAiC20192020 and the Project ID AWI\_PS122\_00. 
    We thank all those who contributed to MOSAiC and made this endeavour 
    possible. \cite{mosaic_ack}
    \bibliographystyle{IEEEtran}
    \bibliography{references}
    
    \appendices
    
    \section{Multimodal Learning for Acoustic-Visual Reconstruction}
    
    The experiments presented in this work demonstrate a complementary relationship between acoustic and visual sensing modalities. Multibeam sonar consistently provides globally stable low-frequency geometry across surveyed regions, while optical reconstruction captures higher-frequency surface detail when visibility, illumination, and feature overlap are sufficient. This asymmetry suggests that future under-ice reconstruction systems should move beyond independent processing pipelines toward mutually constraining multimodal representations of the same environment.
    
    Rather than treating acoustic and visual reconstructions as independent estimates combined only after reconstruction, each modality can inform the reliability and interpretation of the other. Sonar geometry provides a physically grounded structural prior capable of constraining optical reconstruction, while optical signals contribute local detail beyond the spatial resolution limits of acoustic sensing. This motivates a broader interpretation of reconstruction itself: multimodal systems should not only estimate geometry, but also predict where reliable optical reconstruction is expected to succeed.
    
    This perspective is supported by observations made throughout the benchmark experiments. Although the dataset provides broad spatial coverage and multimodal sensor availability, only a subset of regions yields stable Structure-from-Motion (SfM) reconstructions due to strong variability in visibility, texture density, scattering effects, and environmental conditions. In practice, many valid image frames contribute little to successful feature matching or camera pose recovery despite containing usable visual information.
    
    Preliminary experiments conducted outside the scope of the primary benchmark further support this observation. Using approximately 700 unaltered image batches containing 50-100 images each collected from five separate surveys, several lightweight machine learning models were trained to predict whether an Agisoft Metashape SfM reconstruction would be successfully generated based on extracted image statistics, including sharpness, entropy, brightness, keypoint density, and color variance. The best-performing model (CatBoost) achieved an accuracy and weighted F1-score of 0.85 with a ROC-AUC of 0.90. Although limited in scale, these results suggest that the success of optical reconstruction is at least partially predictable from observable photometric and structural image properties.
    
    This naturally motivates future learning-based approaches for reconstruction confidence estimation. A potential extension would involve training a CNN to predict a dense spatial confidence map directly from optical imagery, optionally conditioned on sonar geometry and navigation uncertainty. Regions where SfM or monocular depth estimates agree well with sonar-derived geometry could provide positive supervision examples, while regions exhibiting geometric deviation, fragmentation, or complete reconstruction failure could provide negative examples. Such a framework would allow reconstruction quality to be interpreted as a learnable property of the environment rather than a purely post-hoc evaluation metric.
    
    A potential benefit of this approach is adaptive reconstruction and fusion. Instead of applying SfM or monocular depth estimation uniformly across all regions, confidence estimates could restrict reconstruction to areas where the optical signal is expected to remain reliable, reducing drift in feature-poor or highly scattering environments. Similarly, multimodal fusion could be spatially weighted according to confidence, allowing optical reconstruction to dominate in high-detail regions while sonar geometry remains the primary constraint in visually degraded areas.
    
    \section{Above-Water Ice Reconstruction and Full-Thickness Ice Modelling}
    While this study concentrates on observing the ice from below, it is important to recognize that ice constitutes a single physical entity. Its morphology above the water surface is equally critical for comprehending its mass balance and structural behavior. Drone-based surveys, utilizing RGB cameras and LiDAR, provide an additional method for mapping surface features such as pressure ridges and sails. When these aerial data are precisely aligned with underwater sonar surveys, the separate above- and below-water reconstructions can be combined into a cohesive, full-thickness ice model, simultaneously capturing both the ice draft and its freeboard.
    
    Such a model, encompassing both surfaces, facilitates novel analytical approaches. We can directly compute total ice volume and its thickness distribution with high spatial detail, moving beyond indirect hydrostatic estimates. Moreover, models incorporating complete 3D geometry offer significantly improved predictions regarding internal stress concentrations and the dynamics of deformation. Radiative transfer modeling also profits from this, as the framework inherently connects surface albedo, observed from above, with light transmittance, measured from below.
    
    From a geometric perspective, the above-water surface can act as a guiding input for reconstructing the under-ice environment. This helps narrow down the possible solutions for camera pose estimation in Structure-from-Motion algorithms. In situations where optical methods beneath the ice are ineffective-for instance, in turbid or feature-lacking regions-surface data can stabilize or complete acoustic reconstructions.
    
    Implementing this system necessitates addressing several challenges: managing temporal misalignments caused by ice drift, precisely aligning diverse sensor inputs such as GPS, ROV navigation, and acoustic positioning, and reducing platform instability often induced by wind. Furthermore, the water interface itself represents a significant optical and acoustic discontinuity.
    
    Despite these inherent difficulties, the simulation pipeline presented here provides a logical initial framework. Expanding the HoloOcean environment to incorporate an aerial sensing component will allow for a thorough examination of these co-registration and fusion challenges within a controlled setting, prior to deployment in real-world field operations.
    
    Beyond the geometric information provided by sonar, the variety of shapes within the simulation environment can be significantly increased. This is accomplished using AI-driven 3D mesh generation tools, such as Meshy and Pixal3D. Rather than relying solely on ice structures collected during actual field campaigns, generative mesh pipelines offer the ability to create a wide array of physically realistic ice morphologies on demand. These can range from smooth congelation ice to heavily fractured pressure ridges, and are then directly imported into the HoloOcean environment. This approach substantially broadens the scope of training scenarios without necessitating further field expeditions. Such a capability is especially pertinent for developing reconstruction confidence models. These models require exposure to both structurally simple and geometrically complex ice configurations to achieve robust generalization.
    \begin{figure}[htbp]
    \centering
    \includegraphics[width=\columnwidth]{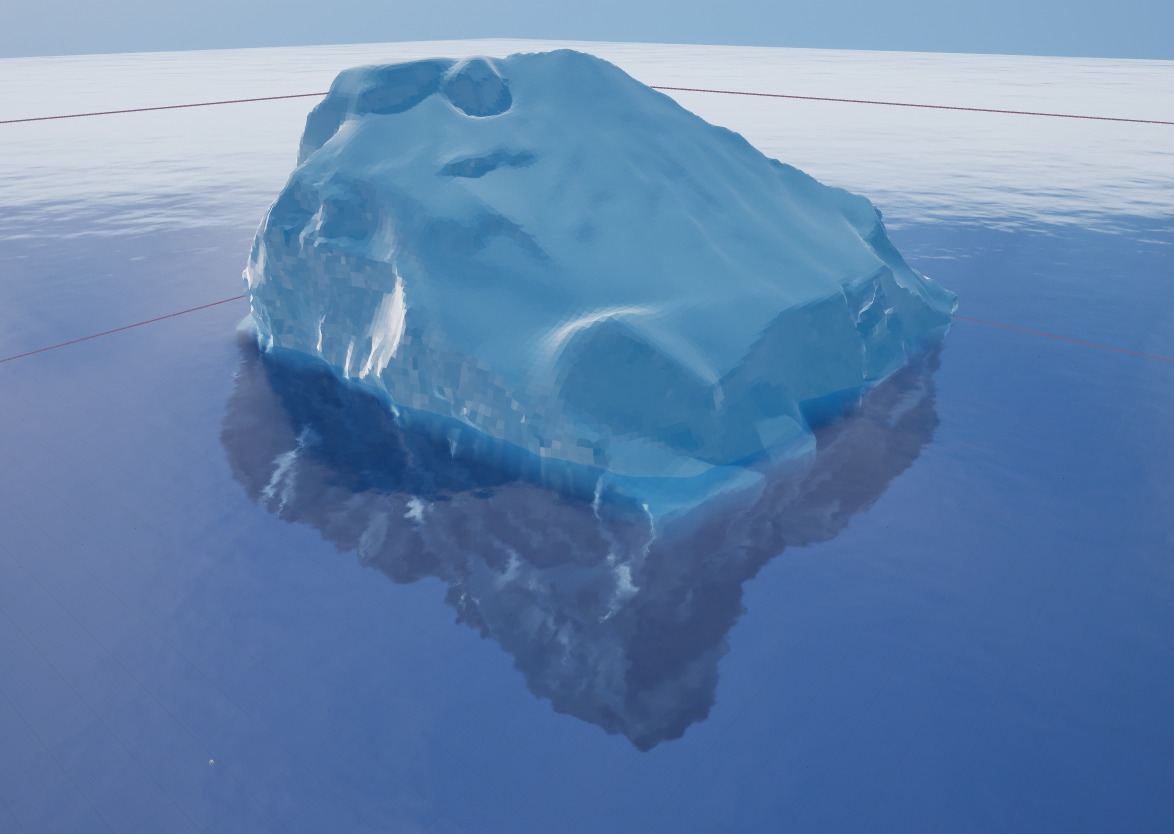}
    \caption{Initial version of a Meshy-generated composite ice structure rendered in HoloOcean, showing irregular above-water geometry and submerged draft. This highlights how generative mesh pipelines can expand morphological diversity in synthetic training datasets without field data collection.}
    \label{fig:meshy_ice}
    \end{figure}
    
    \section{Real-Time and Simultaneous Reconstruction}
    The reconstruction methods evaluated in this benchmark operate offline, processing sensor data after a survey has been completed. Moving towards real-time or simultaneous reconstruction would significantly increase the operational utility of under-ice robotic systems. This would enable capabilities such as adaptive mission planning, on-the-fly obstacle avoidance, and immediate quality assessment of sensor coverage during a survey. Simultaneous Localization and Mapping (SLAM) frameworks, particularly those integrating visual and acoustic cues, represent a natural evolution of the methods presented here. Acoustic SLAM systems have demonstrated considerable robustness in featureless underwater environments, while visual-inertial odometry can provide higher-frequency pose updates where optical conditions allow. A hybrid acoustic-visual SLAM pipeline, wherein sonar constrains global drift and optical flow refines local trajectory estimates, aligns directly with the complementary sensing structure evidenced in the experimental results of this work. The primary limitation for real-time deployment is computational: dense point cloud generation and ICP-based alignment are currently too demanding for onboard embedded hardware at typical survey framerates. Lightweight neural architectures for monocular depth estimation and efficient sparse SLAM backends are active research directions that could help overcome this challenge.
    
    \end{document}